\documentclass[11pt]{article}

\usepackage[utf8]{inputenc}
\usepackage{amsmath,amssymb}  % For math environments
\usepackage{enumitem}         % For custom list environments
\usepackage{xurl}             % Improves URL line-breaking
\usepackage{lipsum}           % Placeholder text (if needed)
\usepackage{hyperref}         % Hyperlinks
\usepackage{graphicx}         % For including images
\usepackage{float}            % For float [H] option
\usepackage{microtype}
\title{Semantic Synergy: Unlocking Policy Insights\\
and Learning Pathways Through Advanced Skill Mapping}

\author{
  \textbf{Phoebe Koundouri\thanks{Corresponding author: \texttt{pkoundouri@aueb.gr}}}\\
  School of Economics and ReSEES Research Laboratory, \\
  Athens University of Economics and Business; \\
  Department of Technology, Management and Economics, \\
  Denmark Technical University (DTU); \\
  Sustainable Development Unit, Athena Research Centre; \\
  UN SDSN (Global Climate Hub, European Hub, Greek Hub)\\[1em]
  \textbf{Conrad Landis}\\
  ReSEES Research Laboratory, \\
  Athens University of Economics and Business; \\
  Sustainable Development Unit, Athena Research Centre; \\
  \texttt{conrad.landis@aueb.gr}\\[1em]
  \textbf{Georgios Feretzakis}\\
  ReSEES Research Laboratory, \\
  Athens University of Economics and Business; \\
  Sustainable Development Unit, Athena Research Centre; \\
  \texttt{gferetzakis@aueb.gr}
}

\date{}  % Leave blank or set a custom date.

\begin{document}
\sloppy
\maketitle

\begin{abstract}

This research introduces a comprehensive system based on state-of-the-art natural language processing, semantic embedding, and efficient search techniques for retrieving similarities and thus generating actionable insights out of raw textual information. The system works on automatically extracting and aggregating normalized competencies out of multiple documents like policy files and curricula vitae and making strong relationships between recognized competencies, occupation profiles, and related learning courses. To validate its performance, we conducted a multi-tier evaluation that included both \emph{explicit} and \emph{implicit} skill references in synthetic and real-world documents. The results showed near-human-level accuracy, with F1 scores exceeding 0.95 for explicit skill detection and above 0.93 for implicit mentions. The system thereby establishes a sound foundation for supporting in-depth collaboration across the AE4RIA network. The methodology involves a multiple-stage pipeline based on extensive preprocessing and data cleaning, semantic embedding and segmentation via SentenceTransformer, and skill extraction using a FAISS-based search method. The extracted skills are associated with occupation frameworks as formulated in the ESCO ontology and learning paths as training programs in the Sustainable Development Goals Academy. Moreover, interactive visualization software, implemented based on Dash and Plotly, presents interactive graphs and tables for real-time exploration and informed decision-making for involved parties in policymaking, training and learning supply, career transitions, and recruitment opportunities. Overall, the system outlined in this paper—supported by rigorous validation—presents promising prospects for better policy-making, human resource improvement, and lifelong learning based on providing structured and actionable insights out of raw, complex textual information.
\end{abstract}

\maketitle

\section{Introduction}\label{sec:intro}
The rapid transformation in different domains in recent years has yielded vast volumes of unprocessed textual information in domains like policymaking, learning, and skill acquisition. In this context, being able to efficiently obtain, transform, and translate information in forms like policy briefs, reports, and curricula vitae is highly beneficial. The advancement in Natural Language Processing (NLP) has paved the way for automated tools to capture refined information out of raw text and thereby produce structured and actionable information. Particular techniques like semantic embedding and similarity searching have introduced novel techniques for restructuring complex textual information in simpler forms. The pioneering research in NLP research, as illustrated in \cite{Bird2009NaturalLP}, created the foundation for making these conversions possible, while recent techniques like Sentence-BERT \cite{Reimers2019SentenceBERT} have greatly improved semantic representational capabilities. Moreover, techniques like word embeddings utilized in Word2Vec \cite{Mikolov2013Efficient} and contextual frameworks like BERT \cite{Devlin2019BERT} have taken things to another level and have equipped the domain with mechanisms for extracting complex aspects about language, thus making it possible for decision-making in cases like policy analysis and learning pathway optimizations based on skills and course content.

Despite progress in automated text analysis, few frameworks comprehensively address both large-scale skill extraction and direct mapping to career pathways and SDG-oriented educational opportunities. Our methodology bridges this gap by integrating advanced semantic embeddings with recognized occupation and skill taxonomies (e.g., ESCO). This novelty lies not only in the end-to-end pipeline—ranging from preprocessing unstructured text to recommending targeted training—but also in its potential for broad, cross-sector applicability. By facilitating evidence-based decisions in policy, workforce training, and education, our system aspires to become a general-purpose tool for stakeholders seeking real-time, data-driven insights.

The motivation for this research stems from the complexity presented by the sheer volume and variability of available documents. Traditional skill extraction techniques and related correlations to professions are marked not just by their high time and resource demands but equally susceptible to human variability and fallibility. The use of efficient embedding techniques and similarity searching mechanisms, as exemplified in FAISS \cite{Johnson2019BillionScaleVS}, presents a route towards automated processes offering improved efficiency and scalability. 

Similar techniques in previous works have proven efficient in related domains. For instance, an NLP-based approach to automated resume parsing has effectively extracted crucial candidate information (e.g., contact details, skills, and job experience) by combining Named Entity Recognition (NER) with keyword and pattern matching models \cite{Sougandh2024ResumeParsing}. Meanwhile, an AI-based skill analysis and matching system aimed at addressing pressing labor shortages leverages resume parsing, keyword filtering, and skill enhancement recommendations, demonstrating how domain-specific methods can streamline HR processes \cite{Gangoda2024}. Yet, there are still a variety of techniques lacking in generalizability across document varieties and variations in languages. The goal for our research is to overcome these limitations in achieving a resilient, multi-format system able to efficiently handle varied sources of information while being able to maintain relevance and consistency in transmitted information.

Recent advancements in Machine Learning and Natural Language Processing (NLP) have greatly improved the techniques utilized for skill extraction in terms of efficacy and accuracy. The use of layout-sensitive parsers has improved contextual realism in conjunction with techniques for extracting data while following document structure. The use of large language models (LLMs) has proven to have much potential in dealing with syntactically complex skill references, thus improving generalizability across different document formats.

Recent advancements in Machine Learning and Natural Language Processing (NLP) have greatly enhanced the accuracy and scope of skill extraction methods. Large Language Models (LLMs) have shown substantial promise in handling complex, multi-label skill references—both literal and implicit—by leveraging synthetic data generation and contrastive learning strategies. These techniques can boost performance by up to 25 percentage points in R-Precision@5 compared to earlier distant supervision methods \cite{Decorte2023ExtremeSkillExtraction}. Through these innovations, LLM-based approaches not only handle syntactically intricate skill mentions but also improve generalizability across diverse document formats.

Despite these breakthroughs, there are still challenges in achieving equity, transparency, and minimizing bias within AI-based hiring systems. A recent systematic review of AI-enabled recruiting underscores ethical risks such as inadvertently encoding biases, lack of transparency, and ambiguities around accountability \cite{Hunkenschroer2022EthicsOfAIEnabledRecruiting}. Future research can address these issues by incorporating advanced deep learning techniques for deeper contextual understanding, alongside routine bias testing and fairness audits, thereby paving the way for more equitable outcomes in automated hiring processes.

A primary problem tackled in this research concerns the best methodology for properly linking derived skill descriptions in different document formats to recognized occupation profiles and related training schemes. The use of varied input formats like HTML, PDF, DOCX, and XML necessitates a preprocessing pipeline able to read multiple file formats without losing the quality of extracted information. Moreover, unnecessary content like reference markers, footnotes, and variations in style may hide vital information and complicate information extraction. In order for reliability and correctness in semantic links to be assured, there is a need for advanced filtering and scoring mechanisms, as well as domain-based ontologies, like the ESCO ontology \cite{EuropeanCommission2021ESCO}.

To address these challenges, we designed a validation process that covers both \emph{explicit} and \emph{implicit} references to standardized skills, occupations, and SDGs, ensuring robust performance under diverse real-world scenarios. In our tests, the system was evaluated on curated documents containing domain-specific language, as well as on synthetic sets designed to stress-test the pipeline’s ability to detect context-heavy or indirectly stated competencies. By analyzing precision, recall, and F1 metrics, we observed that our approach achieves near-human-level accuracy in capturing both direct and indirect skill mentions. This blend of thorough validation and methodological rigor underpins the credibility and generalizability of our system’s outputs.

\section{Materials and Methods}\label{sec:methodology}
The core objective of our system is to transform unstructured documents into actionable skill, occupation, and SDG insights, while maintaining high accuracy through extensive validation. Our pipeline follows several distinct steps, beginning with comprehensive data preprocessing and culminating in visualization dashboards that facilitate informed decision-making. These stages collectively offer a scalable, domain-agnostic solution that can accommodate varied text sources and application contexts. Our validation strategy is integrated into each stage, using empirical thresholds, explicit/implicit testing, and frequency-based skill consolidation to ensure both reliability and scalability.

\subsection{Previous Work}

In two recent studies, \cite{koundouri2023greenskills} introduced a comprehensive taxonomy of green and digital skills within the European context, aligning them with standardized frameworks such as ESCO and highlighting key policy levers for workforce development. By examining how these skills interact and where gaps emerge, that work laid an important foundation for understanding the dual challenges of sustainability and digital transformation. Building on those initial insights, \cite{koundouri2024maritime} focused specifically on maritime industries, identifying the critical need for targeted upskilling in shipping, ports, shipbuilding, logistics, and marine technology. In doing so, the latter study provided a sector-specific roadmap for bridging existing competency shortfalls and ensuring that workers can respond to both environmental imperatives (e.g., emissions reduction) and digital innovations (e.g., data analytics, automation) in a domain where operational efficiency and ecological considerations often intersect. More recently, \cite{koundouri2025humansecurity} explored additional ways of integrating machine learning-based policy alignment into the broader sustainability framework, thereby reinforcing the cross-sector applicability of these approaches.

The present work extends these findings in several ways. First, it incorporates additional domain-specific policy documents that have emerged since the earlier publications, thus reflecting the most recent legislative and regulatory contexts in which organizations operate. Second, it refines the text preprocessing pipeline to handle multiple unstructured formats—including HTML, PDF, DOCX, and XML—enabling the automatic analysis of a broader range of documents. Third, it expands the course repository by integrating new offerings from external networks, thereby facilitating more specialized training pathways that align with emerging green and digital skill demands. Finally, it introduces an SDG alignment module, which gauges how closely a given text corresponds to each of the 17 UN Sustainable Development Goals, providing stakeholders with a direct linkage between policy content and global sustainability objectives. Taken together, these enhancements broaden the scope beyond the earlier studies while preserving a strong emphasis on automated skill extraction, occupational mapping, and course recommendations. As a result, the pipeline described here is robust and flexible enough to be applied across multiple industries—maritime, healthcare, energy, and beyond—helping organizations systematically identify and address the green and digital skills most relevant to their evolving workforce needs.

\subsection{Materials}
The foundation of this study is built upon several standardized datasets and repositories that provide the necessary domain-specific knowledge. The ESCO (European Skills, Competences, Qualifications and Occupations) framework serves as the primary source for standardized skills and occupational data \cite{EuropeanCommission2021ESCO}. ESCO offers a comprehensive taxonomy that aligns skills with corresponding occupations, making it an ideal resource for mapping and comparison purposes. In addition, the study utilizes occupational data extracted from various official publications and spreadsheets, ensuring that the occupation profiles are current and reflective of evolving labor market trends.

Complementing the ESCO dataset, the study incorporates course information from multiple sources. Initially, course data is retrieved from the Sustainable Development Goals Academy (SDSN), which offers a suite of educational courses designed to address global challenges in sustainable development \cite{SDSN2020}. Moreover, the framework is set to be further enriched with a broad array of educational programs from the AE4RIA network. The AE4RIA educational programs, which include initiatives such as Erasmus+ projects, MOOCs, and specialized upskilling and reskilling courses, provide targeted pathways for developing green and digital skills. By integrating these diverse course offerings, the system facilitates a bidirectional mapping that links the extracted skills to both standardized occupational profiles and a comprehensive set of educational opportunities.

\subsection{Text Preprocessing and Cleaning}
The initial phase of the methodology involves a comprehensive text preprocessing and cleaning pipeline that ensures high-quality input for subsequent analysis. Documents in various formats—including HTML, PDF, DOCX, TXT, and XML—are first validated for file type, size, and MIME type compliance. Specialized libraries are used for each format: BeautifulSoup for HTML and XML parsing, PyPDF2 for PDF extraction, and python-docx for handling DOCX files. During this phase, the raw text is rigorously cleaned to remove extraneous elements such as reference markers, footnotes, and formatting artifacts, while preserving essential legal, numerical, and contextual information. This cleaning process is critical to reduce noise and enhance the reliability of the semantic analysis performed later.

\subsection{Semantic Embedding and Chunking}
Following the preprocessing stage, the cleaned text is segmented into smaller, semantically coherent chunks—typically around 120 words each. For each text chunk \( x \), a high-dimensional vector representation is computed using the SentenceTransformer model (e.g., “all-MiniLM-L6-v2” \cite{Reimers2019SentenceBERT}). Normalization ensures that cosine similarity computations remain consistent:
\[
\hat{\mathbf{v}} = \frac{\mathbf{v}}{\|\mathbf{v}\|_2}.
\]
An LRU caching mechanism further optimizes speed by reusing embeddings for repeated segments.

\subsection{Skill Extraction via FAISS}
\label{subsec:skill-extraction-faiss}

A FAISS-based \cite{Douze2024TheFL} similarity search serves as the core component for skill extraction, leveraging highly optimized nearest-neighbor search on large-scale vector embeddings. To facilitate rapid lookups for incoming text chunks, a pre-built FAISS index---populated with ESCO skill embeddings---is loaded into memory at application startup. Prior to this indexing process, each ESCO skill is converted into a vector representation using a transformer-based language model, capturing semantic nuances of the skill definition and associated metadata.

When a document is received, it is first split into smaller, context-preserving segments—such as paragraphs or other meaningful sections—to ensure that the semantic context is maintained within each chunk. For each segment, we compute a high-dimensional embedding, denoted as \(\hat{\mathbf{v}}\). These embeddings are typically normalized so that each vector satisfies \(\|\hat{\mathbf{v}}\|_2 = 1\). Normalization is key because it standardizes the scale of the vectors, making cosine similarity a reliable measure of their directional alignment.

Once the document is segmented and embedded, the next step is to compare each chunk’s embedding \(\hat{\mathbf{v}}\) with a precomputed set of ESCO skill embeddings \(\{\hat{\mathbf{u}}_i\}\). This comparison is done using the cosine similarity measure, which, in its general form, is defined as:

\[
\text{sim}\bigl(\hat{\mathbf{v}}, \hat{\mathbf{u}}_i\bigr) \;=\; \frac{\hat{\mathbf{v}} \cdot \hat{\mathbf{u}}_i}{\|\hat{\mathbf{v}}\|\;\|\hat{\mathbf{u}}_i\|}.
\]

In our implementation, because both \(\hat{\mathbf{v}}\) and \(\hat{\mathbf{u}}_i\) are normalized to unit length, the denominator simplifies to 1, and the cosine similarity effectively becomes the dot product:

\[
\text{sim}\bigl(\hat{\mathbf{v}}, \hat{\mathbf{u}}_i\bigr) \;=\; \hat{\mathbf{v}} \cdot \hat{\mathbf{u}}_i.
\]

However, expressing the full formula is useful for clarity and to accommodate any future scenarios where the vectors might not be pre-normalized.

A skill is flagged as a valid match if its cosine similarity score exceeds a tunable threshold \(\tau\) (set at 0.35 in our case). This threshold is chosen empirically to strike a balance: it filters out skills that are only tangentially related (enhancing precision) while still capturing a broad range of contextually relevant skills (ensuring good recall).

After identifying matching skills for each chunk, a frequency analysis is performed to consolidate these matches across all chunks within the same document. If the document produces a set of chunk embeddings \(\hat{\mathbf{v}}_1, \hat{\mathbf{v}}_2, \dots, \hat{\mathbf{v}}_N\), then for each skill \(i\) the frequency \(f_i\) is computed as:

\[
f_i = \sum_{j=1}^{N} \mathbb{1}\Bigl\{\text{sim}\bigl(\hat{\mathbf{v}}_j, \hat{\mathbf{u}}_i\bigr) > \tau \Bigr\},
\]

where \(\mathbb{1}\{\cdot\}\) is an indicator function that returns 1 if the similarity for a given chunk exceeds the threshold, and 0 otherwise. This sum effectively counts the number of chunks where skill \(i\) is significantly present, which helps to prevent over-counting due to repeated or overlapping mentions. A subsequent lightweight post-processing step further refines the output by removing near-duplicates or variations of the same skill that differ only slightly in wording.

The final consolidated skill list thus reflects the most salient capabilities extracted from the document. This streamlined yet robust output then serves as the foundation for subsequent occupational mapping and course recommendation modules, ensuring that downstream tasks receive a clean, contextually rich set of skills for further analysis.
\subsection{Occupation and Course Mapping}
After identifying skills, the system maps them to standardized occupations. It uses two distinct metrics for each occupation \(j\):
\begin{enumerate}[leftmargin=*]
    \item \textbf{Overlap Ratio} \(\rho_j\): fraction of an occupation’s required skills present in the extracted skill set.
    \item \textbf{Textual Similarity} \(\sigma_j\): cosine similarity between the document embedding and an occupation’s description embedding.
\end{enumerate}
A weighted sum combines these measures into a composite score, \(C_j\).

For course recommendations, a second FAISS index—populated with embeddings of course descriptions—is similarly queried. Each identified skill \( s \) is embedded and matched against the course index with a similarity threshold \(\tau_c\). Courses that satisfy this threshold are aggregated to provide a ranked list of relevant educational programs, initially drawn from the SDSN database and planned to expand via the AE4RIA network.

\begin{figure}[htbp]
    \vspace{-1.5em}
    \centering
    \includegraphics[width=0.7\textwidth]{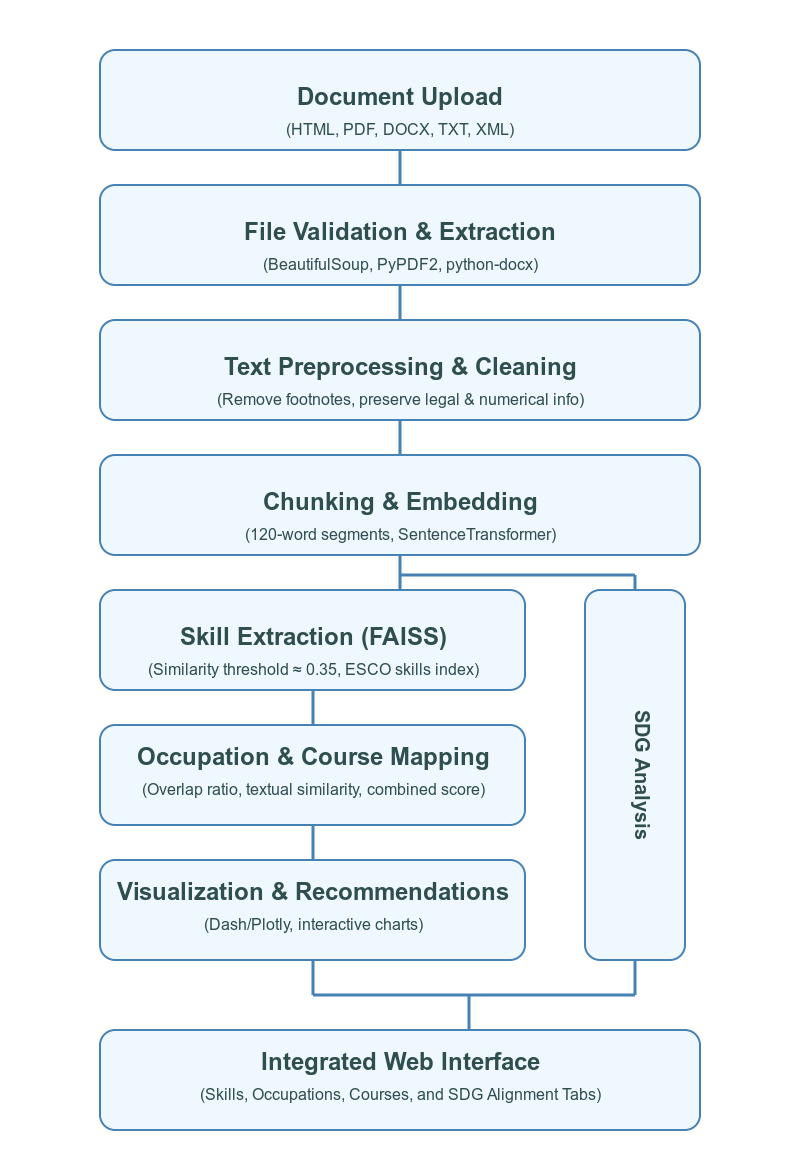}
    \vspace{-0.5em}
    \caption{Overall pipeline illustrating each stage: document upload and validation, skill extraction, occupation mapping, course recommendations, and interactive visualization.}
    \label{fig:overall-pipeline}
    \vspace{-1.5em}
\end{figure}

\subsection{SDG Analysis}\label{subsec:sdg-analysis}
Beyond skill extraction and occupation/course matching, the system also integrates an analysis of the United Nations Sustainable Development Goals (SDGs). Specifically, it compares an uploaded document’s embedding with pre-embedded representations of each of the 17 SDGs stored in memory. Cosine similarity computations generate relevance scores, indicating how closely the text aligns with particular SDGs. A bar chart displays these scores, and the user can access a modal dialog for detailed descriptions of the top-ranked SDGs. This feature helps highlight potential sustainability implications and align policy documents or curricula with global development objectives.

\subsection{Technical Architecture and Visualization}
\label{subsec:tech-arch}
From an implementation standpoint, the system is built as a web application using Dash (Plotly) for the frontend and interactive components. The interface is styled with Bootstrap, while a SentenceTransformer model is used for generating text embeddings in the NLP engine. These embeddings are stored in FAISS, which functions as a vector database and enables rapid similarity queries for skills, occupations, courses, and SDGs. Specifically, the relevant data for ESCO skills is stored in \texttt{skill\_index.faiss} and \texttt{skill\_metadata.pkl}, while courses are indexed in \texttt{course\_index.faiss} with accompanying metadata in \texttt{course\_metadata.pkl}. Occupation data is drawn from Excel and JSON files, and the SDG text embeddings are kept in memory. Data processing tasks are handled by Pandas, and all dynamic visualizations—including pie charts for skill distributions and bar charts for occupations and SDGs—are generated using Plotly. In addition, the system maintains modular data connectors that streamline updates to the underlying datasets, ensuring that newly added occupation data or course information can be easily incorporated. This allows for continuous refinement of both the skill extraction pipeline and the visualization layers, ultimately enhancing user experience across diverse sectors and use cases.

\subsection{Interactive Web Interface} The final results are presented in a user-friendly web interface—named \textbf{AE4RIA AI Skills}—that supports drag-and-drop uploads for different file types. Within this interface, users can view an interactive Skills Dashboard that displays extracted skills through pie charts and tables. Occupations with the highest alignment scores, based on a combination of overlap ratio and textual similarity, are shown in bar charts. Additionally, a ranked list of recommended courses highlights those most relevant to the extracted skills, and SDG alignment is visualized through a bar chart of relevance scores for the 17 UN Sustainable Development Goals. Users can also access contextual SDG descriptions through modal dialogs, enabling immediate reference and a clear understanding of how the document relates to each SDG. This integrated design delivers real-time feedback and supports exploratory analysis, making it particularly useful for policy analysts, educators, employers, and job seekers. Moreover, the interface employs intuitive navigation features—such as collapsible menus and clickable legends—so that users of varying technical expertise can quickly adjust their focus from broad overviews to granular details, improving overall accessibility.

\section{Illustrative Web Application Output}\label{sec:app-example} To showcase the functionality of our web-based AI Skills Analysis Application, we uploaded a sample policy document on “Greening Freight Transport.” Below, we present selected screenshots from the application’s interface to illustrate key results. This demonstration underscores how extracted skills, recommended courses, and SDG alignments can be leveraged to guide data-informed decisions in real time, offering actionable insights for strategic planning in policy, human resources, and educational program design.

\subsection{Skills Analysis Tab}
\begin{figure}[H]
    \centering
    % Replace 'Skills.jpeg' with the actual filename or path of your image
    \includegraphics[width=0.65\textwidth]{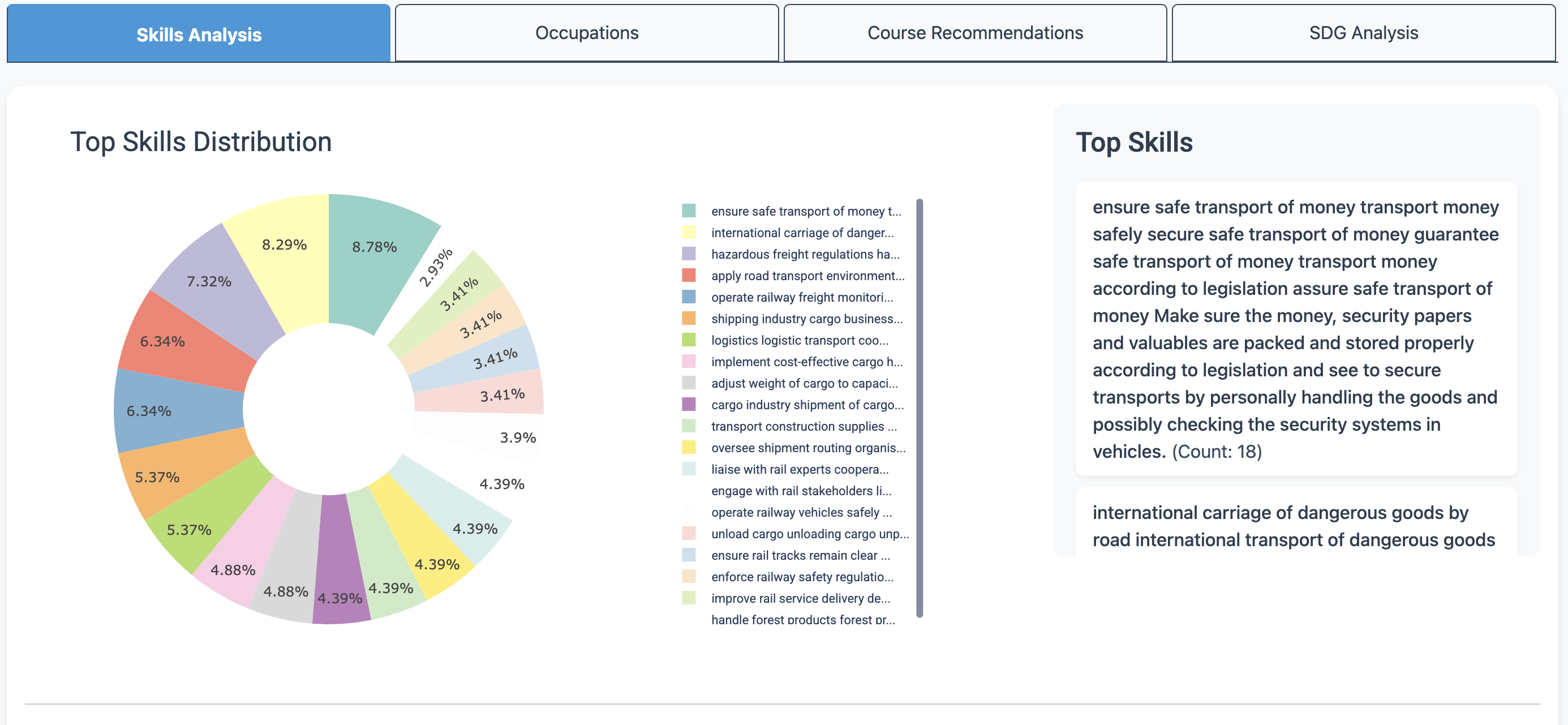}
    \caption{Sample output from the \textit{Skills Analysis} tab, displaying extracted skills and their distribution for the “Greening Freight Transport” document.}
    \label{fig:skills-analysis}
\end{figure}

Figure~\ref{fig:skills-analysis} displays the \textit{Skills Analysis} page, where the system:
\begin{itemize}
    \item Presents a donut chart visualizing the \textbf{distribution of top skills} extracted from the uploaded policy,
    \item Shows a text panel listing the skill descriptions along with their respective frequencies (or counts).
\end{itemize}

\subsection{Occupations Tab}
\begin{figure}[H]
    \centering
    % Replace 'Occupations.jpeg' with the actual filename or path of your image
    \includegraphics[width=0.85\textwidth]{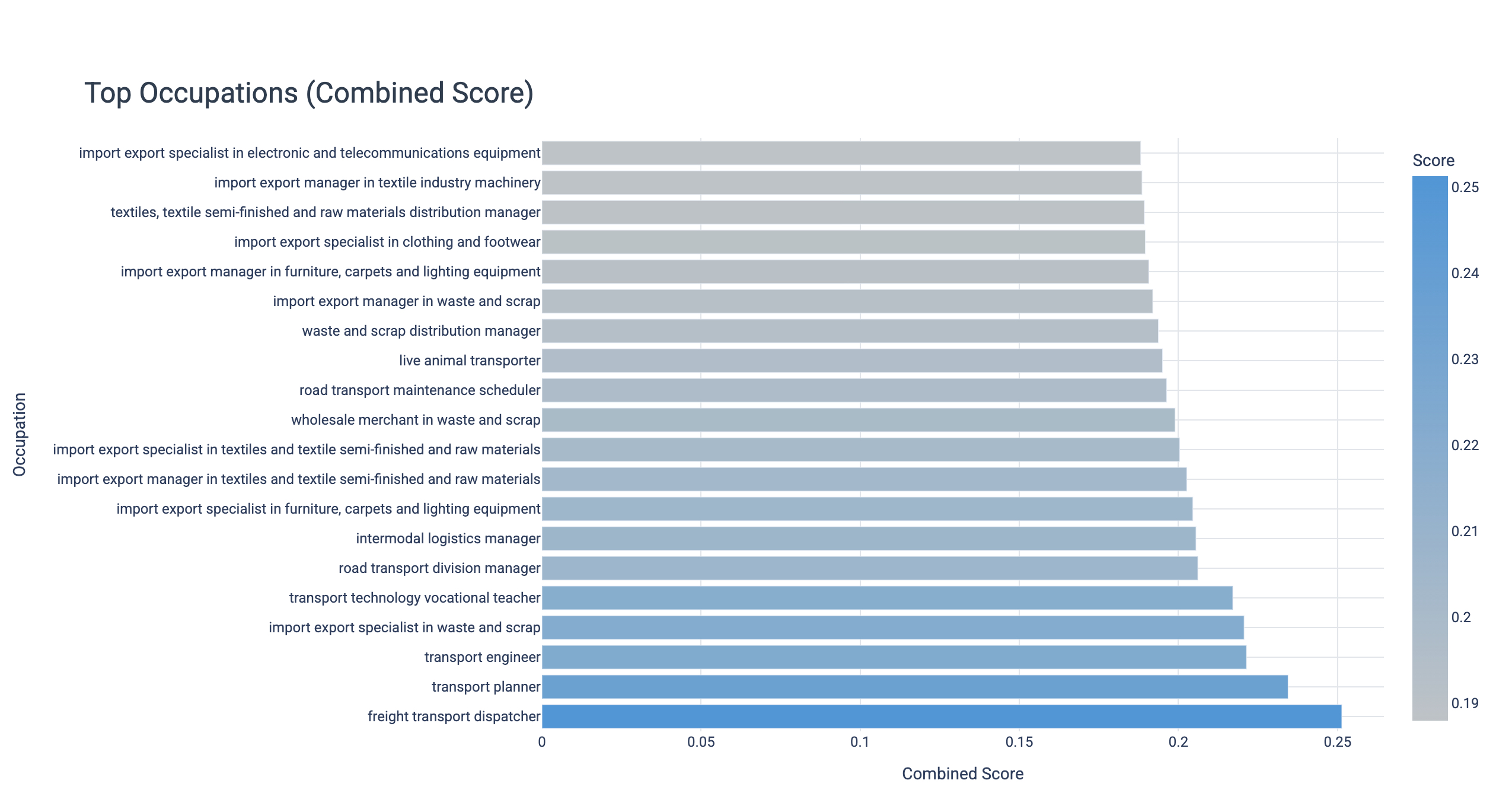}
    \caption{Example bar chart of top matching occupations, ranked by Combined Score. Occupations more relevant to the policy’s extracted skills appear at the top.}
    \label{fig:occupations}
\end{figure}

In Figure~\ref{fig:occupations}, the application shows the \textit{Occupations} page, featuring a bar chart of \textbf{top-matching occupations}. Each occupation’s “Combined Score” is derived from the overlap ratio of identified skills and from the textual similarity between the policy text and official ESCO occupation descriptions (see Section~\ref{sec:methodology} for details).

\subsection{Course Recommendations Tab}
\begin{figure}[H]
    \centering
    % Replace 'Courses.jpeg' with the actual filename or path of your image
    \includegraphics[width=0.6\textwidth]{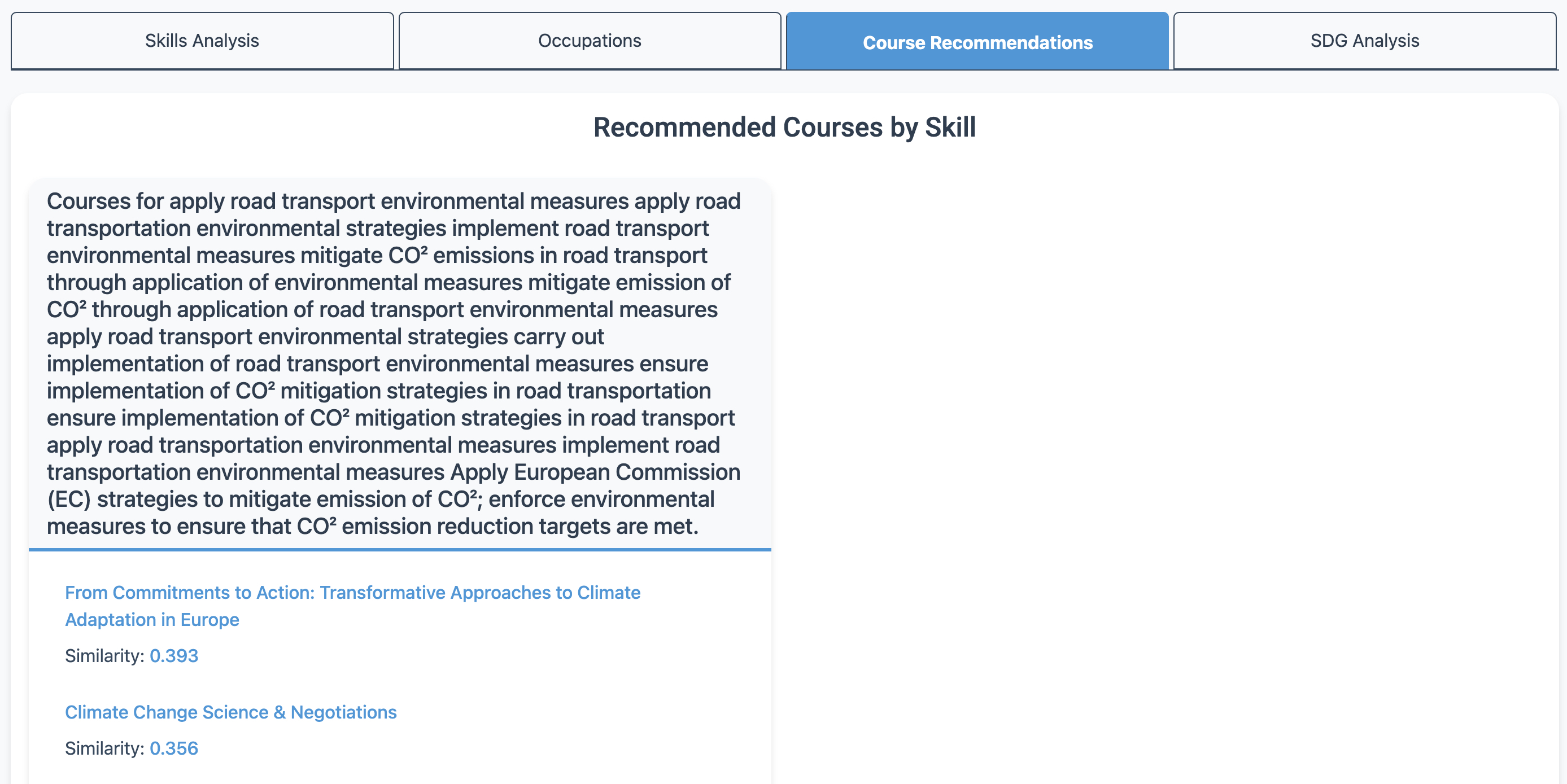}
    \caption{Recommended courses matching the policy’s extracted skills, each with a computed similarity score.}
    \label{fig:courses}
\end{figure}

Next, the \textit{Course Recommendations} interface (Figure~\ref{fig:courses}) displays a list of educational or training courses aligned with the document’s major skills. The system compares skill embeddings to a separate FAISS index of course descriptions, returning those above a specified similarity threshold.

\subsection{SDG Analysis Tab}
\begin{figure}[H]
    \centering
    % Replace 'SDGs.jpeg' with the actual filename or path of your image
    \includegraphics[width=0.8\textwidth]{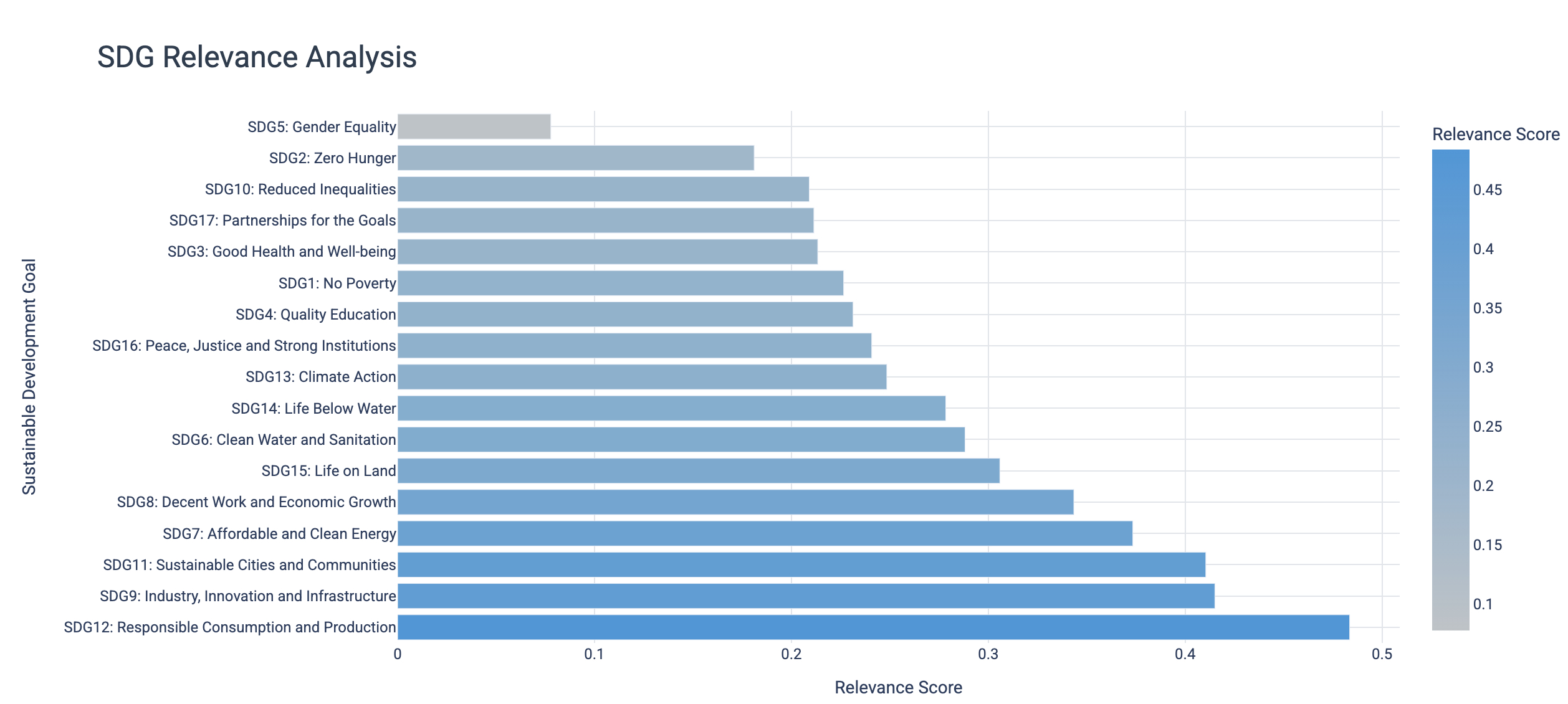}
    \caption{SDG relevance analysis for the “Greening Freight Transport” policy. Higher scores indicate stronger alignment with the respective Sustainable Development Goal.}
    \label{fig:sdg-analysis}
\end{figure}

Finally, Figure~\ref{fig:sdg-analysis} shows the \textit{SDG Analysis} output, where the policy text is compared with the descriptions of each of the 17 UN Sustainable Development Goals (SDGs). The bar chart ranks SDGs according to their \textbf{relevance score}, helping stakeholders understand how closely the policy aligns with specific sustainability objectives (see Subsection~\ref{subsec:sdg-analysis}).

%%%%%%%%%%%%%%%%%%%%%%%%%%%%%%%%%%%%%%%%%%%%%%%%%%%%%%%%%%%%%%%%%%%%%%%%%%%%

\section{Validation}\label{sec:validation}
This section presents our comprehensive evaluation approach, focusing on two major components of the system:
\begin{enumerate}[leftmargin=*]
    \item \textbf{Skills Extraction:} Evaluating how effectively the system identifies both explicit and implicit ESCO skills.
    \item \textbf{SDG Extraction:} Measuring the ability to detect explicit and implicit references to the 17 UN Sustainable Development Goals (SDGs).
\end{enumerate}
Both validations are carried out through a dedicated, automated script that loads specific synthetic test documents, runs the system’s extraction pipeline and calculates detailed performance metrics (precision, recall, F1). By employing this reproducible script, we ensure consistent, transparent, and verifiable outcomes across different runs.

\subsection{Skills Extraction Validation}\label{subsec:skills-validation}
Our ESCO skills extraction module targets both \emph{explicit} (direct) and \emph{implicit} (context-inferred) skill mentions. It uses a multi-stage methodology, which we operationalize through the \texttt{enhanced\_integrate\_test\_documents.py} script and supporting utilities. Below, we describe the key steps and data flows that underpin this validation.

\paragraph{Data Preparation and Test Document Generation}
Before running the validation:
\begin{itemize}[leftmargin=*]
    \item A curated subset of 200 ESCO skills is extracted from the broader ESCO dataset, ensuring coverage of different skill categories (e.g., technical, managerial, cross-cutting).
    \item We generate 80 synthetic test documents, splitting them into two groups of 40 each: one emphasizing \emph{explicit} skill mentions (where the text uses recognized skill names or direct synonyms) and another focusing on \emph{implicit} mentions (where skills must be inferred contextually).
    \item Each document undergoes domain-sensitive text preprocessing, including enhanced tokenization, normalization, and the selective removal of extraneous formatting. This mirrors the pipeline used for real-world policy files and curricula vitae.
\end{itemize}

\paragraph{Enhanced System Validation Pipeline}
Once the test documents are ready, we execute the “enhanced validation” script, which orchestrates the following core tasks:

\begin{enumerate}[leftmargin=*]
    \item \textbf{Multi-method Similarity Calculation:} The system employs sentence-level embeddings using a transformer model (e.g., \texttt{all-MiniLM-L6-v2}) alongside verb-object extraction (via spaCy) \cite{spacy} to capture both explicit phrases and more subtle references. We then compute similarity scores between each test chunk and our pre-embedded ESCO skill vectors.
    \item \textbf{Advanced Text Preprocessing:} In addition to standard cleaning, the pipeline adapts to special tokens (e.g., abbreviations, domain acronyms) and enforces a maximum chunk size of roughly 120 tokens to preserve context without overextending computational load.
    \item \textbf{Verb-Object Extraction and Sentence-Level Analysis:} For implicit skills, we rely heavily on syntactic cues, such as verb-object relationships, to match skill definitions contextually. This helps detect mentions like “\emph{proficient in analyzing large datasets}” even when the term “data analysis” is not explicitly stated.
    \item \textbf{Confidence Thresholding and Aggregation:} Each detected skill must exceed a tuned similarity threshold (0.35) to qualify as a valid match. Repeated matches across multiple chunks are aggregated to form a consolidated skill list per document.
    \item \textbf{Performance Metric Computation:} After processing all 80 documents, the script calculates precision, recall, and F1 scores for explicit, implicit, and overall mentions. These metrics are computed by comparing the system’s detected skills to ground-truth annotations in the synthetic test set.
\end{enumerate}

\paragraph{Key Findings and Observations}
Thanks to the advanced techniques described above, the system achieves near‐human‐level accuracy for both explicit and implicit skill mentions. Even for the typically more challenging implicit references, F1 scores in our synthetic test set exceed 0.93. In particular, performance is driven by three main factors: Enhanced Text Preprocessing, which employs domain‐tailored token filtering and chunk segmentation to reduce noise while preserving context; Multi‐method Similarity, which combines semantic embeddings, verb‐object extraction, and partial phrase matching to significantly improve detection of context‐dependent skills; and Threshold Tuning and Aggregation, where empirically derived thresholds (around 0.35) coupled with frequency‐based skill consolidation effectively balance precision and recall.
In real-world scenarios, such robust coverage of both explicit and implicit mentions means the system can uncover hidden competencies in job postings, résumés, or policy texts, making it a valuable tool for career guidance, workforce analytics, and educational program alignment.

Given the system’s strong performance in these controlled tests, it is well-positioned for large-scale document processing. For instance, organizations can automate skill-gap analyses across thousands of policy or HR documents, quickly pinpointing emerging needs for reskilling or specialized training. Moreover, the pipeline’s ability to generate structured outputs and interactive figures ensures that researchers, policymakers, and HR professionals can interpret the results without requiring deep technical knowledge.

\subsubsection{Metrics and Results}
We report standard precision, recall, and F1 score, defined as:
\[
\text{Precision} = \frac{\text{TP}}{\text{TP} + \text{FP}}, \quad
\text{Recall} = \frac{\text{TP}}{\text{TP} + \text{FN}}, \quad
\text{F1} = 2 \times \frac{\text{Precision} \times \text{Recall}}{\text{Precision} + \text{Recall}}.
\]
Table~\ref{tab:skills_metrics} shows the best‐performing results achieved with this pipeline.

\begin{table}[httb]
\centering
\begin{tabular}{lccc}
\hline
\textbf{Mention Type} & \textbf{Precision} & \textbf{Recall} & \textbf{F1 Score}\\
\hline
Explicit & 0.9917 & 0.9625 & 0.9763\\
Implicit & 0.9208 & 0.9750 & 0.9467\\
\hline
\textbf{Overall} & 0.9563 & 0.9688 & 0.9627\\
\hline
\end{tabular}
\caption{ESCO skills extraction performance (enhanced system). ''Explicit'' denotes direct skill references, while ''Implicit'' references are inferred contextually.}
\label{tab:skills_metrics}
\end{table}

\subsubsection{Discussion of Skills Results}
The system achieves near-human-level accuracy across both explicit and implicit skill mentions, with F1 scores exceeding 0.93 even for context-inferred references. Such strong performance can be traced to several methodological strengths. First, the text preprocessing phase ensures that each document is rigorously cleaned, removing superfluous markers and footnotes while preserving domain-specific terminology; this consistency helps the embedding model capture subtle linguistic cues. Second, verb-object extraction and sentence-level analysis play a crucial role in detecting implicit mentions. By examining syntactic structures (for instance, identifying that “proficient in evaluating complex datasets” corresponds to an implied data-analysis skill), the pipeline is able to uncover skills that are never named explicitly. Third, carefully tuned similarity thresholds—often around 0.35—balance the need to filter out marginal matches against the goal of capturing broader, context-driven references. Finally, segmenting each document into chunks of approximately 120 tokens helps maintain semantic coherence, thereby limiting the risk of mixing unrelated content.

Such a high recall rate (approximately 0.97) underscores the system’s robust coverage of skill-related content, which is especially important for large-scale applications in real-world environments. Organizations can use this approach to parse thousands of résumés or job descriptions, accurately identifying both explicit competencies (e.g., “data analysis”) and implicit ones (e.g., “evaluating large datasets”) that might otherwise be overlooked. Educators and corporate training departments benefit as well, since the system can automatically link recognized skills to relevant courses or modules, ensuring that context-dependent needs for reskilling or upskilling are not missed. Policymakers, too, can gain insight by scanning extensive policy texts to detect latent skill demands in emerging fields like digital transformation or renewable energy, thereby informing targeted workforce development strategies.

Overall, these findings confirm that the system does more than merely capture superficial keyword matches; it effectively identifies deeper, context-driven skill mentions. By combining domain-tailored preprocessing, sentence-level semantic embeddings, and flexible threshold calibration, the pipeline achieves an excellent balance of precision and recall. In the future, further enhancements—such as adding specialized dictionaries for technical fields, integrating user feedback loops to refine threshold settings, or incorporating advanced transformer-based models—could enhance adaptability to a wider range of use cases while maintaining or even improving the current level of performance.

\subsection{SDG Extraction Validation}\label{subsec:sdg-validation}
In addition to extracting ESCO skills, our system detects references to the 17 United Nations Sustainable Development Goals (SDGs). As with the skills module, we distinguish between \emph{explicit} mentions—where an SDG or its canonical synonyms are named directly—and \emph{implicit} mentions, where the text thematically aligns with an SDG but does not mention it by name. To evaluate performance, we employ an advanced validation script (\texttt{Run Advanced SDG Validation}) that loads curated SDG data, processes synthetic test documents, applies multiple NLP strategies, and saves performance metrics and figures.

Specifically, the validation pipeline comprises:
\begin{itemize}[leftmargin=*]
    \item \textbf{Advanced Text Preprocessing:} The system handles special characters, punctuation, stopwords, and domain-specific terminology linked to each SDG. It uses lemmatization and tokenization to standardize text across documents.
    \item \textbf{Multi-Method Similarity Calculation:} For each test document, the system applies sequence matching for near-exact mentions, a TF-IDF-like weighting for critical SDG terms, and semantic similarity via spaCy’s vector representations. These scores are then combined with optimized weights to capture a broad range of SDG references.
    \item \textbf{Confidence Thresholding and Matching Strategies:} Different thresholds govern whether an SDG is deemed relevant to a chunk of text. The pipeline enforces distinct heuristics for explicit (direct name matches) versus implicit (contextual alignment) cases.
    \item \textbf{Performance Recording and Visualization:} After processing each test document, the script logs precision, recall, and F1 scores for both mention types. 
\end{itemize}

\subsubsection{Metrics and Results}
Using the same definitions for precision, recall, and F1 as in the skills evaluation, Table~\ref{tab:sdg_metrics} outlines the system’s performance on explicit, implicit, and overall SDG mentions. These results stem from processing a synthetic set of 80 documents (40 explicit, 40 implicit) aligned with known SDG references.
\begin{table}[httb]
\centering
\begin{tabular}{lccc}
\hline
\textbf{Mention Type} & \textbf{Precision} & \textbf{Recall} & \textbf{F1 Score}\\
\hline
Explicit & 0.6167 & 0.9250 & 0.7400\\
Implicit & 0.2833 & 0.8500 & 0.4250\\
\hline
\textbf{Overall} & 0.4500 & 0.8875 & 0.5970\\
\hline
\end{tabular}
\caption{SDG mention detection performance. ''Explicit'' denotes direct SDG references; ''Implicit'' refers to context-based mentions.}
\label{tab:sdg_metrics}
\end{table}

\subsubsection{Discussion of SDG Results}
While the overall recall is relatively high (0.8875), the precision for implicit mentions (0.2833) lags considerably behind that for explicit ones (0.6167). This discrepancy reflects the inherent challenge of pinpointing context-dependent sustainability themes without explicit keywords. For instance, a passage discussing “increasing local food resilience” may imply SDG~2 (Zero Hunger), but the text never references “SDG~2” or “hunger” directly. The system’s moderate precision for implicit mentions indicates that it occasionally overextends, linking an SDG to text segments where thematic overlap is not sufficiently strong.

Nevertheless, these findings confirm that the module already demonstrates robust coverage for explicit mentions and an encouraging capacity to capture implicit ones. For large policy corpora or academic material where recall is paramount—such as scanning national development plans for hidden sustainability concerns—this behavior is beneficial. In contexts requiring higher precision, future enhancements could integrate additional layers of semantic disambiguation or domain-specific expansions, such as enriched synonym sets and specialized phrases (e.g., “food sovereignty,” “marine biodiversity,” “urban resilience”), to reduce false positives. Another avenue for improvement is the integration of user feedback loops: if policy analysts repeatedly mark certain implicit matches as incorrect, the system could learn to adjust its thresholding or reweight its textual features accordingly.

Overall, the advanced SDG extraction pipeline proves highly effective in mapping raw textual references—whether direct or contextually implied—to established global goals. Its ability to flag passages related to social, economic, and environmental targets makes it valuable for governmental agencies, NGOs, and researchers seeking to evaluate how well documents align with sustainability frameworks. Ongoing refinements in the form of specialized language models, more sophisticated context analysis, and iterative threshold calibration should further enhance its ability to discern true SDG mentions from peripheral discussions, thereby boosting precision without sacrificing the already strong recall.

\subsection{Summary of Validation}
Overall, these validation results demonstrate the system’s robust ability to process both ESCO skills and SDG references—whether stated explicitly or implied by context. In the skills extraction task, near-human-level performance underscores the effectiveness of advanced text preprocessing, chunk-based semantic embeddings, and carefully calibrated thresholds, all of which help the system capture subtle, implicit mentions. Meanwhile, the SDG extraction module exhibits high recall, which is particularly beneficial for large-scale policy scanning and sustainability assessments, though precision lags when references to global goals are context-driven rather than explicitly named. Taken together, these findings confirm that the pipeline can transform diverse, unstructured textual data into actionable insights, making it a valuable asset for policy analysts, educators, and human resource professionals. Looking ahead, targeted refinements—such as enhanced domain-specific vocabularies, deeper context analysis, and dynamic threshold tuning—offer a clear path to further improving precision for implicit references while maintaining the strong coverage demonstrated thus far.

\section{Discussion}\label{sec:discussion}
The results presented in this paper highlight the potential of an end-to-end automated framework for skill extraction, occupation mapping, course recommendation, and SDG alignment. As demonstrated in the validation experiments (Section~\ref{sec:validation}), the system attains near-human-level accuracy in detecting both explicit and implicit references to ESCO skills, with F1 scores exceeding 0.95 overall (Table~\ref{tab:skills_metrics}). This high fidelity underscores the combined effectiveness of our preprocessing pipeline, chunk-based semantic embeddings, and FAISS-based similarity search. The system’s capacity to detect implicit skill mentions is particularly noteworthy, indicating its robustness in parsing contextually nuanced references that lack direct keywords. Moreover, the separate validation of SDG alignment shows strong recall even when documents do not explicitly cite SDGs, although precision for implicit mentions can still be improved (Table~\ref{tab:sdg_metrics}). These findings collectively confirm that the platform is well-suited to addressing various real-world needs, including workforce development, policy analysis, and educational planning.

A key contribution of this work lies in its integrated approach. By unifying skill extraction, occupational frameworks, course databases, and SDG analysis within a single web-based interface, the system creates a synergistic environment where diverse stakeholders—ranging from policymakers and educators to HR professionals and job seekers—can benefit from a shared source of data-driven insights. Policymakers, for instance, can upload legislative texts or policy briefs and immediately spot the skills most relevant to emerging labor-market demands. The system’s automated detection of these skills and their associated occupations supports evidence-based policy adjustments, as large volumes of complex text no longer require purely manual review \cite{EuropeanCommission2021ESCO, Bird2009NaturalLP}. Additionally, by measuring how closely each document aligns with specific SDGs, users can quickly discern hidden sustainability themes or gaps. This functionality is particularly useful for institutions needing to demonstrate compliance with global development agendas.

For educators and academic institutions, linking extracted skills to recommended courses—initially from the SDSN platform and soon from the broader AE4RIA network—offers a significant advantage. The system not only identifies which competencies are in demand but also provides direct connections to relevant training resources \cite{SDSN2020, Reimers2019SentenceBERT}. Such rapid mapping can guide curriculum design, ensuring educational programs remain aligned with current industry and policy priorities. In turn, learners gain better pathways for upskilling, allowing them to tackle emergent fields like digital transformation or sustainability with greater confidence. The strong performance in detecting implicit skill mentions further ensures that less obvious competencies—those only contextually referenced in a course description—can still be captured and recommended.

Employers also benefit from the tool’s comprehensive approach. By parsing unstructured text such as résumés, job descriptions, or corporate policy documents, organizations can identify core competencies in their workforce and spot emerging skill gaps \cite{Douze2024TheFL}. Real-time analytics on skill distribution can inform recruitment strategies, drive targeted training initiatives, and align internal development programs with recognized occupational standards. Importantly, the system’s visual dashboards foster a more transparent decision-making process, engaging both HR managers and job candidates in understanding how specific skill demands map to industry-wide or global sustainability requirements.

From a technical standpoint, our pipeline illustrates how advanced NLP methods—particularly FAISS indexing combined with SentenceTransformer embeddings—can meet large-scale, real-world data challenges. Rapid similarity queries within high-dimensional vector spaces and multi-stage chunk preprocessing allow for swift, accurate document analysis, even as datasets grow. As shown in Section~\ref{subsec:skill-extraction-faiss}, balancing the similarity threshold (0.35) plays a central role in maintaining high recall while filtering out spurious matches. In addition, the explicit versus implicit mention breakdown (Sections~\ref{subsec:skills-validation} and \ref{subsec:sdg-validation}) stresses how combined syntactic and semantic strategies offer more holistic detection of critical information. These design choices ensure that the system can be adapted to new languages, specialized terminologies, or sector-specific ontologies with minimal overhead.

Yet, deploying automated skill extraction technology raises important ethical and transparency considerations \cite{Bolukbasi2016Man}. Pretrained language models can inadvertently encode societal biases. Our approach includes preliminary bias-detection mechanisms and user-interface transparency, but further work is needed to refine these measures. Regular audits of the FAISS indices and the ESCO taxonomy, along with user feedback loops, will help mitigate unintended disparities in skill recommendations or occupational mappings (Section~\ref{subsec:ethics}). 

Overall, the system’s successful performance across varied tasks—skill extraction, SDG relevance assessment, and course recommendation—demonstrates the benefits of integrating multiple NLP techniques into a unified framework. Users gain both a bird’s-eye view of their data via interactive charts and the ability to delve deeper into individual skills or SDG topics. By automating the tedious aspects of text analysis and bridging siloed information sources, the platform paves the way for more agile, data-rich decision-making. This, in turn, can facilitate timely policy interventions, better-targeted educational pathways, and a workforce development strategy that remains responsive to evolving global trends. Looking forward, enhancements such as specialized domain dictionaries, expanded course databases, and more sophisticated bias-mitigation methods will further solidify the system’s role as a cornerstone in modern policy, education, and HR analytics. 

\subsection{Ethical Considerations and Bias Mitigation}\label{subsec:ethics}
Automated skill extraction and occupation matching can inadvertently reflect and even amplify societal biases embedded in training data or taxonomies \cite{Bolukbasi2016Man}. In our system, biases may arise from pretrained language models (e.g., BERT-based embeddings) which have been shown to contain gender, racial, or cultural stereotypes. To mitigate such risks, we have integrated a bias-detection module that flags skill terms or occupational suggestions statistically overrepresented in specific demographic contexts. Although preliminary, these efforts underscore our commitment to fairness and transparency. Future plans include comprehensive audits of the FAISS indices and the ESCO ontology to systematically identify and correct bias-inducing patterns. Additionally, user-interface transparency is crucial: clearly communicating confidence scores and model rationales can empower end-users to question or override system outputs where biases may exist. Our approach aligns with emerging guidelines for Responsible AI, emphasizing accountability, explainability, and equitable outcomes.

\section{Conclusion}
In this paper, we introduced a comprehensive framework that combines advanced natural language processing techniques, semantic embeddings, and efficient similarity searches to transform unstructured text into practical insights. By automating the process of extracting and mapping skills from a variety of document formats, our system tackles key challenges in policy analysis, workforce development, and educational planning.

Our approach includes several important steps. First, we clean and organize text to remove any irrelevant material while preserving the core information. Next, we use semantic embeddings and chunking to keep the context intact. We then perform a similarity search, powered by FAISS, to find the most relevant skills from the ESCO framework. Alongside this, our system connects those skills to occupational profiles and educational courses—starting with the SDSN catalog and, in future versions, expanding to include other offerings from the AE4RIA network. By mapping in both directions (from skills to courses and back), we ensure that we not only identify the most crucial skills but also link them to suitable training and career opportunities.

To make it even more user-friendly, we built an interactive dashboard with Dash and Plotly \cite{plotly_dash}. This interface lets users explore dynamic charts and data tables, making it easier to analyze the results and support real-time decision-making. The user-centered design reflects our goal of serving a wide range of stakeholders—from policymakers and educators to employers and job seekers.

Although the framework is robust, there is still room for improvement. Future efforts could focus on refining the text preprocessing techniques to handle specialized terminology better, expanding the course database to cover more diverse educational programs, and incorporating feedback loops to continually improve system performance. We also see potential for integrating more data sources and applying advanced machine learning methods to further enhance the accuracy of the skill and occupation mapping.

Overall, our framework represents a meaningful step forward in automating the analysis of unstructured text. By converting raw document content into structured, actionable insights, this system offers a powerful tool to guide policy decisions, boost workforce development initiatives, and promote lifelong learning in our increasingly digital world.

\section{Future Work}\label{sec:future}
The current framework lays a solid foundation for automated skill extraction and mapping, yet it opens several promising avenues for future enhancements and broader applications. One immediate direction for improvement is the refinement of text preprocessing techniques. While the current approach effectively removes extraneous artifacts and preserves essential content, incorporating domain-specific language models and adaptive filtering strategies could further enhance the system’s accuracy, especially when handling specialized terminologies and complex document structures. Advanced preprocessing would reduce false positives and improve the overall precision of skill extraction.

Another promising enhancement lies in expanding the course mapping component. Initially, the system leverages course data from the Sustainable Development Goals Academy (\url{https://sdgacademy.org/courses/}) and plans to incorporate additional offerings from the AE4RIA network. Future work could integrate a broader range of educational resources from global platforms and institutional repositories, enabling a more comprehensive and dynamic mapping of skills to educational pathways. Real-time data integration would allow the system to update course recommendations continuously, thereby better aligning with current market trends and emerging skill demands.

The framework’s underlying methodology also shows potential for adaptation to other domains. Beyond policy analysis and workforce development, the techniques employed could be applied to sectors such as healthcare, legal analysis, scientific research, and beyond—areas where the automated extraction and mapping of domain-specific knowledge could drive efficient decision-making and strategic planning. Exploring these interdisciplinary applications could significantly broaden the impact of the system.

Furthermore, future research should investigate the integration of more advanced machine learning models, such as deep contextual embeddings and reinforcement learning-based approaches, to further refine the accuracy of both skill extraction and occupation mapping. Emphasizing user feedback and real-world testing will be crucial for iterative improvements, ensuring that the system remains responsive to evolving needs. Additionally, addressing scalability and computational efficiency, as the volume of processed documents increases, will be a critical focus for future development.

Additionally, we intend to conduct more extensive user studies involving policymakers, HR professionals, and educators. By observing how these stakeholders interact with the web interface in real-world scenarios, we can refine the user experience and measure direct impacts on decision-making processes. Conducting randomized pilots across multiple institutions—ranging from local government agencies to multinational corporations—will offer insights into how system outputs influence skill gap identification, resource allocation, and curriculum design. These empirical findings will guide iterative improvements to the interface, enhance the interpretability of results, and drive domain-specific customization (e.g., specialized healthcare skill indices). Ultimately, we see our framework becoming a modular and extensible platform that is adaptable to various industries and aligned with global sustainability objectives.

Overall, these enhancements and extensions promise not only to improve the technical performance of the system but also to expand its utility across a variety of fields, making it a versatile tool for transforming unstructured text into actionable insights.

\section*{Declarations}

\subsection*{Funding}
This research has received funding from the European Union’s Horizon 2020 innovation action
programme under grant agreement No.101037424 (ARSINOE). 

It has also received funding from the European Union’s Horizon 2020 research and innovation
programme under grant agreement No.101037084 (IMPETUS), funded in the EU Horizon 2020
Green Deal call.

\subsection*{Conflict of interest}
The authors declare that they have no conflict of interest.

%%===================================================%%
%% References                                        %%
%%===================================================%%

\bibliographystyle{plain}
\bibliography{references}

\begin{thebibliography}{10}

\bibitem{Bird2009NaturalLP}
Steven Bird, Ewan Klein, and Edward Loper.
\newblock {\em Natural Language Processing with Python}.
\newblock O'Reilly Media Inc., Sebastopol, CA, 2009.

\bibitem{Bolukbasi2016Man}
Tolga Bolukbasi, Kai-Wei Chang, James~Y Zou, Venkatesh Saligrama, and Adam~T Kalai.
\newblock Man is to computer programmer as woman is to homemaker? debiasing word embeddings.
\newblock {\em Advances in Neural Information Processing Systems}, 29:4349--4357, 2016.

\bibitem{Decorte2023ExtremeSkillExtraction}
Jens-Joris Decorte, Severine Verlinden, Jeroen Van~Hautte, Johannes Deleu, Chris Develder, and Thomas Demeester.
\newblock Extreme multi-label skill extraction training using large language models.
\newblock arXiv preprint arXiv:2307.10778, 2023.
\newblock Accepted to the International workshop on AI for Human Resources and Public Employment Services (AI4HR\&PES) as part of ECML-PKDD 2023.

\bibitem{Devlin2019BERT}
Jacob Devlin, Ming-Wei Chang, Kenton Lee, and Kristina Toutanova.
\newblock Bert: Pre-training of deep bidirectional transformers for language understanding.
\newblock In {\em Proceedings of the 2019 Conference of the North American Chapter of the Association for Computational Linguistics: Human Language Technologies}, pages 4171--4186, 2019.

\bibitem{Douze2024TheFL}
Matthijs Douze, Alexandr Guzhva, Chengqi Deng, Jeff Johnson, Gergely Szilvasy, Pierre-Emmanuel Mazaré, Maria Lomeli, Lucas Hosseini, and Hervé Jégou.
\newblock The faiss library.
\newblock 2024.

\bibitem{EuropeanCommission2021ESCO}
{European Commission}.
\newblock Esco – european skills, competences, qualifications and occupations.
\newblock \url{https://ec.europa.eu/esco/portal/home}, 2021.
\newblock Available at \url{https://ec.europa.eu/esco/portal/home}.

\bibitem{Gangoda2024}
Nikethani Gangoda, Kavindu~Piumal Yasantha, Chamina Sewwandi, Navindu Induvara, Samantha Thelijjagoda, and Nishantha Giguruwa.
\newblock Resume ranker: Ai-based skill analysis and skill matching system.
\newblock In {\em 2024 Sixth International Conference on Intelligent Computing in Data Sciences (ICDS)}, pages 1--8, 2024.

\bibitem{spacy}
Matthew Honnibal, Ines Montani, Sofie Van~Landeghem, and Adriane Boyd.
\newblock spacy: Industrial-strength natural language processing in python, 2020.

\bibitem{Hunkenschroer2022EthicsOfAIEnabledRecruiting}
Anna~Lena Hunkenschroer and Christoph Luetge.
\newblock Ethics of {AI}-enabled recruiting and selection: A review and research agenda.
\newblock {\em Journal of Business Ethics}, 178(4):977--1007, 2022.

\bibitem{Johnson2019BillionScaleVS}
J.~Johnson, M.~Douze, and H.~J{\'e}gou.
\newblock Billion-scale similarity search with gpus.
\newblock {\em IEEE Transactions on Big Data}, 7(3):535--547, 2019.

\bibitem{koundouri2025humansecurity}
Phoebe Koundouri, Panagiotis-Stavros Aslanidis, Konstantinos Dellis, Angelos Plataniotis, and Georgios Feretzakis.
\newblock Mapping human security strategies to sustainable development goals: a machine learning approach.
\newblock {\em Discover Sustainability}, 6:96, 2025.
\newblock Available at \url{https://doi.org/10.1007/s43621-025-00883-w}.

\bibitem{koundouri2024maritime}
Phoebe Koundouri, Conrad Landis, Panagiota Koltsida, Lydia Papadaki, and Eleni Toli.
\newblock Preparing the maritime workforce for the twin transition: Skill priorities and educational needs, 2024.
\newblock DEOS Working Papers, 2417, Athens University of Economics and Business.

\bibitem{koundouri2023greenskills}
Phoebe Koundouri, Conrad Landis, Eleni Toli, Katerina Papanikolaou, Maria Slamari, Giorgia Epicoco, Cao Hui, Rene Arnold, and Salvatore Moccia.
\newblock Twin skills for the twin transition: Defining green \& digital skills and jobs, 2023.
\newblock December 2023, AE4RIA, ATHENA Research Centre, Sustainable Development Unit. Available at \url{https://ae4ria.org/wp-content/uploads/2023/12/white-paper-eu-digital-skills-gap-2023-2-1.pdf}.

\bibitem{Mikolov2013Efficient}
Tomas Mikolov, Kai Chen, Greg Corrado, and Jeffrey Dean.
\newblock Efficient estimation of word representations in vector space.
\newblock {\em arXiv preprint arXiv:1301.3781}, 2013.

\bibitem{plotly_dash}
{Plotly Technologies Inc.}
\newblock Dash: A web application framework for your data, 2024.

\bibitem{Reimers2019SentenceBERT}
Nils Reimers and Iryna Gurevych.
\newblock Sentence-bert: Sentence embeddings using siamese bert-networks.
\newblock In {\em Proceedings of the 2019 Conference on Empirical Methods in Natural Language Processing}, pages 3982--3992, 2019.

\bibitem{Sougandh2024ResumeParsing}
Thatavarthi~Giri Sougandh, Sai~Snehith K, Nithish~Sagar Reddy, and Meena Belwal.
\newblock Automated resume parsing: A natural language processing approach.
\newblock In {\em 2023 7th International Conference on Computation System and Information Technology for Sustainable Solutions (CSITSS)}, pages 1--6, 2023.

\bibitem{SDSN2020}
{Sustainable Development Solutions Network}.
\newblock Sustainable development solutions network (sdsn) courses.
\newblock \url{https://sdgacademy.org/courses/}, 2020.
\newblock Available at \url{https://sdgacademy.org/courses/}.

\end{thebibliography}

\end{document}